\documentclass{article}

\usepackage[main, final]{neurips_2026}
\usepackage{multirow}
\usepackage{adjustbox}
\newcommand{\rothead}[1]{\rotatebox{90}{#1}}
\usepackage{xspace}
\usepackage[table]{xcolor}
\usepackage{amsmath}
\usepackage{fvextra}

\newcommand{\ours}{GraphWrit3R\xspace}
\definecolor{lightgreen}{RGB}{220, 250, 220}
\definecolor{jsonbg}{RGB}{245,246,248}
\definecolor{jsonframe}{RGB}{245,246,248}

\usepackage{pifont}
\definecolor{mygreen}{RGB}{0,150,0}
\definecolor{darkorangeyellow}{RGB}{204,130,0}
\newcommand{\cmark}{\textcolor{mygreen}{\ding{51}}}
\newcommand{\xmark}{\textcolor{red}{\ding{55}}}

\usepackage[utf8]{inputenc} %
\usepackage[T1]{fontenc}    %
\usepackage{url}            %
\usepackage{booktabs}       %
\usepackage{amsfonts}       %
\usepackage{nicefrac}       %
\usepackage{microtype}      %
\usepackage{xcolor}         %

\usepackage{caption}
\usepackage{enumitem}

\definecolor{darkred}{rgb}{0.6, 0.1, 0.05}

\usepackage[pagebackref=true,breaklinks=true,colorlinks,bookmarks=false]{hyperref}
\usepackage{cleveref}

\definecolor{mycitecolor}{HTML}{195a66}
\hypersetup{
  citecolor  = mycitecolor,
  colorlinks = true,
}

\newcommand{\PAR}[1]{\vskip4pt \noindent{\bf #1~}}

\title{\ours: End-to-End 3D Scene Graph Writing}

\author{%
  Luka Milivojevic\textsuperscript{1}
  \quad Nikola Popovic\textsuperscript{1}\textsuperscript{$\dagger$}
  \quad Sayan Deb Sarkar\textsuperscript{2}
  \quad Sebastian Koch\textsuperscript{3}
  \\
  \textbf{Iro Armeni\textsuperscript{2}}
  \quad
  \textbf{Luc Van Gool\textsuperscript{1}}
  \quad
  \textbf{Danda Pani Paudel\textsuperscript{1}}
  \\[2mm]
  \textsuperscript{1}INSAIT, Sofia University ``St. Kliment Ohridski''
  \\
  \textsuperscript{2}Stanford University
  \qquad
  \textsuperscript{3}Ulm University
}

\begin{document}

\maketitle

\renewcommand{\thefootnote}{}%
\footnotetext{$\dagger$ Project lead}%
\renewcommand{\thefootnote}{\arabic{footnote}}

\begin{center}
    \vspace{-7mm}
    \includegraphics[width=0.93\linewidth]{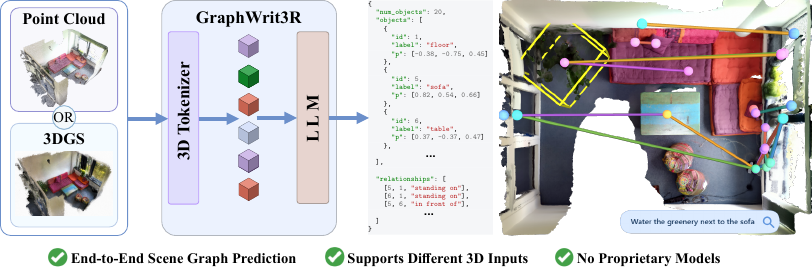}
    \captionof{figure}{
    \textbf{\ours: End-to-End 3D Scene Graph Writing.} \ours takes a point cloud, 3D Gaussian Splats, or both as input and directly generates a scene graph as a structured JSON script. Unlike prior multi-stage pipelines with fragile intermediate representations, objects and their relationships are jointly predicted within a single end-to-end latent model, without reliance on proprietary models. The resulting graph additionally supports open-vocabulary querying.
    }
    \label{fig:teaser}
\end{center}

\begin{abstract}
\vspace{-1mm}
\label{sec:abs}
3D scene graphs provide a structured representation of complex environments by encoding objects, their semantic attributes, and the spatial and functional relationships between them. Current approaches for 3D scene graph generation suffer from several fundamental limitations. They rely on complex multi-stage pipelines with explicit intermediate representations, making systems fragile and prone to error propagation. They assume access to ground-truth object annotations during inference, which deviates from real-world scenarios. They depend on proprietary models, hindering open-source deployment, or incur prohibitively slow inference. We present \ours, a simple end-to-end method that takes a 3D point cloud, Gaussian Splats, or a combination of both as input, and directly outputs a complete scene graph as a structured JSON script. The graph lists all objects, their semantic attributes, and the relationships between them, while avoiding all of the above mentioned limitations. The choice of multiple input modalities is purely for versatility, allowing a single set of weights to handle diverse scenarios. Point cloud inputs are encoded via Sonata and Gaussian Splat inputs via Chorus, with both modalities projected onto a shared voxel grid and fused through a novel per-voxel contrastive alignment loss before being decoded by a large language model. As a natural consequence of the LLM, \ours also supports open-vocabulary querying. On the 3DSSG benchmark, our method achieves state-of-the-art performance on object class, predicate, and triplet recall, outperforming methods that rely on ground-truth object annotations during inference. We further provide qualitative results and analyze different input modality configurations, contrastive loss formulations, and token fusion strategies. %
Our project page is available at \href{https://graphwrit3r.insait.ai}{graphwrit3r.insait.ai}.
  
\end{abstract}

\section{Introduction}

The community has made tremendous progress in 3D scene understanding through semantic segmentation~\cite{wu2025sonata,zhang2025concerto,zhang2026utonia}, object detection~\cite{V-DETR,rukhovich2022fcaf3d,lazarow2024cubify}, instance decomposition~\cite{Schult23,yin2024sai3d,wu2024ptv3}, and visual-language grounding~\cite{lee2025mosaic3d,li2025SceneSplat7k,li2025chorus}. However, these advances have largely focused on individual components rather than the relational structure that connects them. 3D scene graphs~\cite{armeni20193d,wald2020learning,hughes2022hydra} directly address this gap, offering a structured representation that encodes objects, their semantic attributes, and the spatial and functional relationships that tie them into a coherent whole. %
In robotic manipulation, grasping a target object that is \emph{under} another object first requires moving the object \emph{on top of} it out of the way. The robot must resolve spatial relations between objects to determine the correct action sequence. 
In AR, anchoring virtual content in a physical room requires reasoning about which surfaces support objects, which items occlude others, and how a user query like ``show me everything on the shelf'' maps onto the actual 3D arrangement in a view-consistent way. 
Nevertheless, scene graphs remain significantly underexplored compared to their instance-level counterparts. 

Current approaches for 3D scene graph generation suffer from several fundamental limitations. First, most methods rely on complex multi-stage pipelines~\cite{koch2024open3dsg,koch2025relationfield,xie2026relags,linok2025beyond,zhang2025open,gu2024conceptgraphs,Fu_2026_funfact,rotondi2025fungraph}, where objects are detected or segmented from 3D and/or 2D inputs, per-object features are extracted, and graph nodes and edges are initialized via multi-view feature aggregation before being classified by a graph neural network, frozen LLM, or VLM. Each stage introduces its own failure modes, making the overall system fragile, prone to error propagation, and computationally inefficient. Second, many existing methods assume access to ground-truth object bounding boxes or instance masks during inference~\cite{wald2020learning,zhang2021exploiting,koch2024lang3dsg,koch2024sgrec3d,koch2024open3dsg,koch2025relationfield,xie2026relags,linok2025beyond} an assumption that is rarely satisfied in real-world scenarios where such annotations are unavailable. Any object or mask estimator used as a substitute inevitably introduces additional errors that cascade through the remainder of the pipeline. Third, many recent methods rely heavily on proprietary models via API calls~\cite{linok2025beyond,zhang2025open,gu2024conceptgraphs,Fu_2026_funfact,rotondi2025fungraph}, making them unsuitable for full open-source deployment. Finally, methods that avoid proprietary dependencies often suffer from prohibitively slow inference, requiring minutes or hours to process a single scene~\cite{koch2024open3dsg,xie2026relags,linok2025beyond,koch2025relationfield}.

\begin{figure}[t]
    \centering
    \includegraphics[width=0.9\textwidth]{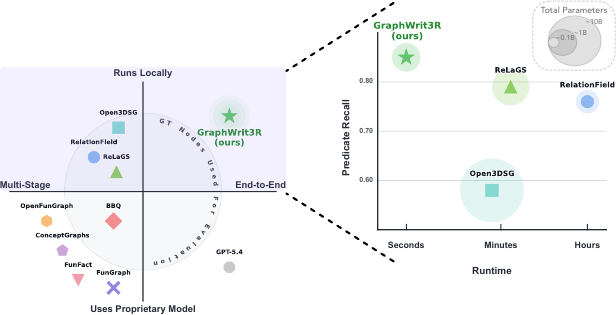}
    \caption{
    \textbf{Comparison to Relevant Methods.} On the left, we categorize relevant scene graph generation methods along several key dimensions. \ours is the only method operating in a fully end-to-end latent fashion, requiring no fragile intermediate explicit representations, no proprietary models calls, and no ground-truth object annotations at inference time. On the right, we zoom in on methods that do not rely on proprietary models, showing that \ours simultaneously achieves the highest performance and orders of magnitude faster inference.}
    \vspace{-3mm}
    \label{fig:relevant_methods}
\end{figure}

We introduce \ours to address these limitations. Given a 3D scene as a point cloud, or a set of Gaussian Splats~\cite{kerbl3Dgaussians}, or a multi-modal combination of both, \ours directly generates a scene graph as a structured JSON script, listing all objects, their semantic attributes, and the relationships between them (see Fig.~\ref{fig:teaser}). As a natural consequence of the LLM backbone, it also supports open-vocabulary scene graph querying. \ours operates entirely in latent space, requiring no fragile intermediate explicit representations, no proprietary model calls, and no ground-truth object annotations as input during inference (see Fig.~\ref{fig:relevant_methods}). \textit{The choice to support multiple input modalities is motivated purely by versatility}, rather than performance. We train with multi-modal inputs and evaluate with a single modality during inference, reflecting real-world settings where different 3D representations exist across different environments. Point cloud inputs are encoded via Sonata~\cite{wu2025sonata} into spatially grounded tokens, while Gaussian Splat inputs are encoded via Chorus~\cite{li2025chorus}, both producing representations on a shared voxel grid. When both modalities are present, tokens at co-located voxels are fused and passed to the Qwen2.5-0.5B~\cite{qwen2025qwen25technicalreport} LLM, which decodes the full scene graph. To enable coherent cross-modal fusion, we introduce a per-voxel contrastive alignment loss that brings Sonata and Chorus features into a common representational space.

Our method is inspired by SpatialLM~\cite{SpatialLM}, which encodes point clouds with Sonata and uses an LLM to produce a structured JSON list of objects from indoor scenes. 
We leverage this foundation model by initializing both our Sonata encoder and LLM from the SpatialLM indoor fine-tuning checkpoint, providing a strong 3D prior. 
However, \ours departs from SpatialLM, in both scope and design. Rather than predicting a flat list of objects, we extend the output representation to a full relational scene graph. Also, we introduce a complementary Gaussian Splat encoding branch via Chorus. Combined with a voxel-level fusion mechanism trained with a contrastive alignment objective, this yields a system with one set of weights that handles multiple input modalities. Finally, we fine-tune the resulting architecture end-to-end for scene graph generation.

We evaluate \ours on the 3DSSG benchmark~\cite{wald2020learning} across standard recall-based metrics for object class, predicate, and relationship triplet prediction, achieving state-of-the-art results against established baselines. Notably, most competing methods operate under the privileged assumption of ground-truth object annotations at inference time and incur substantially higher inference latency. We further analyze different input modality configurations, contrastive loss formulations, and token fusion strategies. Finally, we demonstrate that jointly supervising object detection and relationship prediction yields stronger detection performance than object supervision alone, suggesting that relational context provides a useful inductive signal for grounding individual objects.

Our contributions can be summarized as follows: 
\begin{itemize}[itemsep=1pt, parsep=0pt, topsep=0pt] %
    \item We propose an end-to-end scene graph generation method outputting structured JSON, without intermediate explicit representations, ground-truth annotations, or proprietary models. %
    \item We introduce a unified multi-modal 3D encoder accepting point clouds, or 3DGS, or both, with a novel per-voxel contrastive alignment loss for cross-modal fusion.
    \item We demonstrate state-of-the-art scene graph results on 3DSSG, analyze key design choices, and show that relational supervision benefits object detection.
\end{itemize}

\section{Related work}
\label{sec:relatedwork}

\noindent \textbf{3D Perception.} Early 3D scene understanding relied on complex multi-stage pipelines involving explicit reconstruction, proposal generation, and feature aggregation. Driven by advancements in 3D backbones~\cite{qi2016pointnet,qi2017pointnet++,choy20194d,yang2023swin3d,Zhao_2021_ICCV,wu2022point,wu2024ptv3}, the field has largely shifted toward end-to-end latent-space architectures. This transition is evident in object detection, where early pipeline-heavy methods~\cite{qi2019deep,obj-dgcnn,detr3d} have been superseded by fully end-to-end detectors~\cite{V-DETR,rukhovich2022fcaf3d,lazarow2024cubify}. Similarly, in segmentation, multi-stage clustering and per-scene optimization approaches~\cite{jiang2020pointgroup,nguyen2023open3dis,lu2023ovir,takmaz2023openmask3d,Peng2023OpenScene,knaebel2026ditr,koch2025unified} are rapidly being replaced by unified latent architectures for both closed-~\cite{yang2023swin3d,Zhao_2021_ICCV,wu2022point,wu2024ptv3,wu2025sonata,zhang2025concerto,zhang2026utonia} and open-vocabulary~\cite{lee2025mosaic3d,li2025SceneSplat7k,ma2025scenesplatpp,li2025chorus} settings. Despite this broader trend, 3D scene graph generation remains stuck in the multi-stage paradigm. Existing methods~\cite{wald2020learning, koch2024sgrec3d, wu2021scenegraphfusion, koch2024lang3dsg, koch2024open3dsg, wang2023vl, gu2024conceptgraphs, linok2025beyond} first extract explicit instances to build an initial graph, then apply GNNs~\cite{wald2020learning, koch2024sgrec3d, wu2021scenegraphfusion, koch2024lang3dsg, wang2023vl} or LLMs~\cite{gu2024conceptgraphs, linok2025beyond, koch2024open3dsg} to predict node and edge properties. This cascaded design introduces complexity and propagates errors. To our knowledge, \ours is the first method to break this pattern. By utilizing a fully end-to-end latent pipeline, our approach directly generates a structured scene graph in JSON format without relying on explicit intermediate predictions.

\noindent \textbf{3D Scene Graphs.} Beyond object-centric perception, 3D scene graphs model inter-object relationships for scene understanding. Early works introduced hierarchical representations~\cite{armeni20193d} and formalized semantic relationships~\cite{wald2020learning}. Subsequent methods enabled real-time estimation~\cite{hughes2022hydra} and refined predictions via incremental estimation, scene priors, and pre-training~\cite{wu2021scenegraphfusion,wu2023incremental,zhang2021exploiting,koch2024sgrec3d,koch2024lang3dsg,wang2023vl}. More recent approaches further improve closed-vocabulary prediction through edge-centric relational reasoning~\cite{ma2026leo} and volumetric multi-granularity feature modeling for objects and their relations~\cite{huang2026granssg}. However, these approaches remain confined to fixed, training-time taxonomies. Recent methods achieve open-vocabulary capabilities by integrating foundation models. One line of work distills features into language-queryable 3D representations using GNNs~\cite{koch2024open3dsg}, radiance fields~\cite{koch2025relationfield,xie2026relags}, or hierarchical graphs~\cite{werby2024hovsg}. Recent work also explores geometry-guided discrete diffusion to jointly reason over object geometry and semantic relationships~\cite{feng2026geode}. Another line of work extracts relationships by aggregating natural-language outputs from frozen LLMs~\cite{gu2024conceptgraphs,linok2025beyond}, often using advanced prompting for fine-grained affordances~\cite{zhang2025open,rotondi2025fungraph}. Furthermore, recent work extends open-vocabulary scene graphs toward language-guided interaction and downstream tasks by combining dynamic graph construction with retrieval-augmented reasoning~\cite{yu2026openworld}. Despite their flexibility, these methods rely on multi-stage pipelines and stacked, costly API calls to closed-source models~\cite{achiam2023gpt,comanici2025gemini}, significantly increasing latency. In contrast, we formulate 3D scene graph generation as an efficient, end-to-end learning objective. By removing dependencies on external APIs, our approach enables efficient and local predictions that generalize  across diverse environments.

\noindent \textbf{3D Vision-Language Models.} 3D vision-language models (3D VLMs) enable natural language interaction with 3D scenes. Following early 3D captioning pipelines~\cite{chen2023end}, modern architectures generally encode 2D and 3D modalities into tokens for LLM-driven visual question answering (VQA). These models employ diverse encoding strategies, including joint fine-tuning of 3D object tokens~\cite{wang2023chat}, cross-attention with frozen encoders~\cite{chen2023ll3da}, and multi-modal fusion of 2D, 3D, or BEV features~\cite{huang2024chat,huang2024embodied,Fu_2025_WACV,zhou2023uni3d}. Recent advances further refine this via hierarchical tokenization~\cite{li20243dmit,yu2025inst3d}, adaptive resolution~\cite{zhi2024lscenellm}, or Gaussian Splat inputs~\cite{halacheva2025gaussianvlm}. Despite these architectural advances, most 3D VLMs produce plain-text VQA answers lacking spatial grounding, limiting their utility in embodied applications. While recent models like SceneScript~\cite{avetisyan2024scenescript} and SpatialLM~\cite{SpatialLM} address this by outputting flat JSON lists of detected 3D objects, they ignore inter-object relationships. 
Our \ours goes a step further by extending structured language outputs to the relational structure of the scene, moving beyond flat object lists toward a richer, relationally grounded representation of 3D environments.

\section{Method}
\label{sec:method}

\subsection{Overview}
\label{subsec:overview}

\begin{figure}[t]
    \centering
    \includegraphics[width=\textwidth]{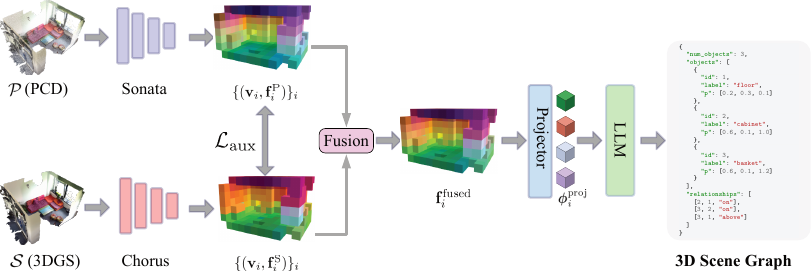}
    \caption{\textbf{\ours.} Each 3D modality is processed by a designated encoder. Sonata~\cite{wu2025sonata} encodes point clouds and Chorus~\cite{li2025chorus} encodes Gaussian splats into voxel tokens on a shared grid. Tokens at co-located voxels are averaged, projected by an MLP into the LLM's embedding space, and decoded autoregressively into a scene graph in structured JSON format.}
    \label{fig:sgllm_pipeline}
    \vspace{-3mm}
\end{figure}

\ours predicts a 3D scene graph
$\mathcal{G} = (\mathcal{O}, \mathcal{R})$ as a structured JSON, given a 3D scene represented as a point cloud $\mathcal{P}$, or Gaussian Splatts $\mathcal{S}$, or both. Here,
$\mathcal{O} = \{o_i\}_{i=1}^{N}$ denotes the set of detected objects, where each object
$o_i = (\ell_i, \mathbf{p}_i, \mathbf{s}_i, \theta_i)$ contains a natural-language semantic label
$\ell_i$, 3D centroid $\mathbf{p}_i \in \mathbb{R}^3$, bounding box extents $\mathbf{s}_i \in \mathbb{R}^3$,
and yaw angle $\theta_i \in \mathbb{R}$. The relation set
$\mathcal{R} = \{\{(o_i, r_{ij}^{(m)}, o_j)\}_{m=1}^{M_{ij}}\}_{i,j=1}^N$ contains directed subject-predicate-object triplets,
where $r_{ij}^{(m)}$ is a natural-language predicate and $M_{ij}$ denotes the number of predicates between objects $o_i$ and $o_j$. Multiple relations may exist between the same pair of objects. Fig.~\ref{fig:sgllm_pipeline} shows the full pipeline. Sonata~\cite{wu2025sonata} encodes $\mathcal{P}$ and Chorus~\cite{li2025chorus} encodes $\mathcal{S}$ into tokens on a shared voxel grid (Sec.~\ref{subsec:encoding}). Co-located voxel tokens are averaged when both modalities are present, or passed through otherwise. A two-layer MLP projects the fused tokens into the embedding space of the LLM, which decodes $\mathcal{G}$ autoregressively as a JSON script (Sec.~\ref{subsec:sgprediction}). Training combines token-level cross-entropy loss at the LLM output  with a contrastive per-voxel alignment loss between Sonata and Chorus (Sec.~\ref{subsec:training_pipeline}). Furthermore, a single set of learned weights for the whole architecture handles all three input configurations at inference, namely $\mathcal{P}$, $\mathcal{S}$, or both. %
Therefore, Sonata and Chorus must produce tokens in a shared latent space, which is enforced by our alignment loss as explained in Sec.~\ref{subsec:training_pipeline}. Moreover, we build on SpatialLM~\cite{SpatialLM} as a foundation model, rather than training a 3D LLM from scratch, and reuse its Sonata encoder, projector, and LLM weights at the beginning of our training.

\subsection{Scene Graph Generation}
\label{subsec:sgprediction}
\label{subsec:encoding}

\PAR{Point Cloud Encoding.} Given a point cloud $\mathbf{P} \in \mathbb{R}^{N_p \times 6}$ of $\{x, y, z\}$ coordinates and RGB color, Sonata~\cite{wu2025sonata} returns voxelized tokens $\mathbf{Z}_{\mathrm{P}} = \{(\mathbf{v}_i,\mathbf{f}_i^{\mathrm{P}})\}_{i=1}^{M_p}$,
where $\mathbf{v}_i \in \mathbb{Z}^3$ is a sparse voxel coordinate, $\mathbf{f}_i^{\mathrm{P}} \in \mathbb{R}^d$ is its feature, and $M_p$ is the number of occupied voxels. Sonata is a sparse voxel transformer built on the PTv3-M2 backbone~\cite{wu2024ptv3}, pretrained by self-distillation on 140k indoor and outdoor scenes. 

Its key design choice is to prevent the \emph{geometric shortcut}, where point coordinates leak into the attention operator and cause features to collapse onto surface geometry. Sonata mitigates this by obscuring spatial information at coarser scales, forcing the network to rely on input features such as color instead.
We initialize Sonata from the fine-tuned checkpoint of SpatialLM~\cite{SpatialLM}, which was fine-tuned jointly with the downstream MLP and LLM for indoor scene parsing.

\PAR{Gaussian Splats Encoding.} Given Gaussian Splat $\mathcal{S} = \{(\boldsymbol{\mu}_j, \boldsymbol{\Sigma}_j, \mathbf{c}_j, \alpha_j)\}_{j=1}^{N_s}$, where for each splat $j$, $\boldsymbol{\mu}_j \in \mathbb{R}^3$ denotes the center, 
$\boldsymbol{\Sigma}_j \in \mathbb{R}^{3 \times 3}$ the covariance matrix parameterized via scale and rotation~\cite{kerbl3Dgaussians}, 
$\mathbf{c}_j \in \mathbb{R}^3$ the color, and $\alpha_j \in \mathbb{R}$ the opacity, Chorus~\cite{li2025chorus} outputs  $\mathbf{Z}_{\mathrm{S}} = \{(\mathbf{v}_i,\mathbf{f}_i^{\mathrm{S}})\}_{i=1}^{M_s}$,
where $\mathbf{v}_i \in \mathbb{Z}^3$ is a sparse voxel coordinate, $\mathbf{f}_i^{\mathrm{S}} \in \mathbb{R}^d$ is its feature, and $M_s$ is the number of occupied voxels that are on the same voxel grid as Sonata. Chorus also uses the PTv3-M2 backbone, making it a feed-forward sparse voxel transformer with the same structure as Sonata. 
The key difference lies in pretraining. Rather than self-supervision, Chorus distills three 2D foundation models into a shared backbone via teacher-specific projection heads, namely SigLIP2~\cite{tschannen2025siglip} for language alignment, DINOv3~\cite{simeoni2025dinov3} for visual features, and PE-Spatial~\cite{bolya2025perception} for object awareness.
The shared PTv3 backbone between the two encoders is what makes feature-level alignment between them tractable.

\PAR{Fusion.} When both 3D modalities are present, we average features at every voxel location $\mathbf{v}_i$ from both encoders. 
\begin{equation}
\mathbf{f}_i^{\mathrm{fused}} = \tfrac{1}{2}\left(\mathbf{f}_i^{\mathrm{P}} + \mathbf{f}_i^{\mathrm{S}}\right), \qquad i \in \mathcal{V}_{\mathrm{P}} \cap \mathcal{V}_{\mathrm{S}}.
\end{equation}
If a specific voxel of one modality does not have a counterpart feature in the other, which happens infrequently due to the different nature of $\mathcal{P}$ and $\mathcal{S}$, that voxel is discarded. If only one modality is used during inference, $\mathbf{Z}_{\mathrm{fused}}$ reduces to either $\mathbf{Z}_{\mathrm{P}}$ or $\mathbf{Z}_{\mathrm{S}}$.

\PAR{Projection.} A two-layer MLP $\phi$ maps each fused token $\mathbf{f}_i^{\mathrm{fused}} \in \mathbb{R}^d$ into the LLM's input embedding space. We initialize $\phi$ from the projector of SpatialLM and fine-tune it end-to-end.

\PAR{LLM.} The projected tokens $\{\boldsymbol{\phi}_i^{\mathrm{proj}}=\phi(\mathbf{f}_i^{\mathrm{fused}})\}_{i=1}^{M}$, $\boldsymbol{\phi}_i^{\mathrm{proj}} \in \mathbb{R}^h$
enter the LLM after a fixed text prompt that specifies the desired output structure. Following
SpatialLM~\cite{SpatialLM}, the LLM autoregressively decodes a structured JSON script
representing the scene graph $\mathcal{G}$ defined in Sec. ~\ref{subsec:overview}. The sequence
contains object entries with semantic labels, 3D centroids, box extents, and yaw angles, followed
by relationship entries represented as subject-predicate-object triplets.

\subsection{Training Objective}
\label{subsec:training_pipeline}

\PAR{Cross-Modal Alignment.} Since Sonata and Chorus produce tokens on the same voxel grid (Sec.~\ref{subsec:encoding}), we can pair their features by voxel coordinate. For every dual-modality scene in a batch, let
\begin{equation}
\mathcal{B} = \{(\mathbf{f}_i^{\mathrm{P}}, \mathbf{f}_i^{\mathrm{S}}) : \mathbf{v}_i \in \mathcal{V}_{\mathrm{P}} \cap \mathcal{V}_{\mathrm{S}}\}
\end{equation}
be the set of co-located token pairs, and let $\tilde{\mathbf{f}} = \mathbf{f} / \|\mathbf{f}\|_2$ denote $\ell_2$-normalization. The cross-modal alignment loss has three terms, each acting on these pairs. 
The first is symmetric InfoNCE~\cite{rusak2024infonce}, where the positive for a Sonata token at voxel $\mathbf{v}_i$ is the Chorus token at the same voxel, and the negatives are Chorus tokens at all other voxels within the same scene. Denoting the set of co-located voxel locations in that scene by $\mathcal{V}$:
\begin{equation}
\mathcal{L}_{\mathrm{InfoNCE}} = -\frac{1}{2|\mathcal{V}|}\sum_{i \in \mathcal{V}} \left[\log \frac{\exp(s_{ii}/\tau)}{\sum_{j \in \mathcal{V}} \exp(s_{ij}/\tau)} + \log \frac{\exp(s_{ii}/\tau)}{\sum_{j \in \mathcal{V}} \exp(s_{ji}/\tau)}\right],
\end{equation}
with $s_{ij} = \langle \tilde{\mathbf{f}}_i^{\mathrm{P}}, \tilde{\mathbf{f}}_j^{\mathrm{S}}\rangle$ and temperature $\tau$. The loss is computed independently for each scene and averaged across the batch.
The second is a per-pair cosine term $\mathcal{L}_{\mathrm{cos}} = 1 - \langle \tilde{\mathbf{f}}_i^{\mathrm{P}}, \tilde{\mathbf{f}}_i^{\mathrm{S}}\rangle$, and the third is a feature-level MSE term $\mathcal{L}_{\mathrm{mse}} = \|\mathbf{f}_i^{\mathrm{P}} - \mathbf{f}_i^{\mathrm{S}}\|_2^2$. The three terms do different things: cosine pulls paired tokens onto a shared direction, MSE pulls them onto a shared magnitude, and InfoNCE keeps tokens at different voxels apart, so that alignment does not collapse the feature space. The combined alignment loss is: $\mathcal{L}_{\mathrm{aux}} = \lambda_{\mathrm{nce}}\mathcal{L}_{\mathrm{InfoNCE}} + \lambda_{\mathrm{cos}}\mathcal{L}_{\mathrm{cos}} + \lambda_{\mathrm{mse}}\mathcal{L}_{\mathrm{mse}}$.
The features of Sonata are already aligned with the LLM in SpatialLM, since they were jointly trained. By contrast, Chorus, was pretrained to match 2D foundation model features. %
Therefore, we stop the propagation of $\mathcal{L}_{\mathrm{aux}}$ gradients through Sonata, such that they propagate exclusively through Chorus. This pulls Chorus's feature space closer to Sonata's, which keeps the LLM's input distribution stable and lets the cross-entropy signal continue to land where SpatialLM expects it. The shared PTv3 backbone between the two encoders is what makes the asymmetric alignment well-posed. If the architectures differed, matching feature statistics would not imply matching feature semantics.

\PAR{Final Loss.}
We supervise the JSON output with the standard next-token cross-entropy loss $\mathcal{L}_{\mathrm{CE}}$, and add the auxiliary alignment loss with scalar weight $\alpha$ to form the full objective $\mathcal{L} = \mathcal{L}_{\mathrm{CE}} + \alpha\,\mathcal{L}_{\mathrm{aux}}$. Each training batch contains a mix of point-cloud-only, Gaussian-only, and dual-modality scenes, while $\mathcal{L}_{\mathrm{aux}}$ is computed only on the dual-modality subset. This yields a single set of weights that handles all three input configurations at inference, namely $\mathcal{P}$, $\mathcal{S}$, or both.

\section{Experiments}
\label{sec:exps}
\subsection{Experimental Setup}

\PAR{Datasets and tasks.}
We evaluate scene graph prediction on 3DSSG~\cite{wald2020learning}, using its evaluation vocabulary of 160 object classes and 27 predicate classes. For training, we use $1,170$ point-cloud scenes from SceneVerse~\cite{jia2024sceneverse}, of which $620$ have Gaussian-splat representations from SceneSplat++~\cite{ma2025scenesplatpp}, with scene graph annotations from 3DSSG. For out-of-domain evaluation, we additionally use the official ScanNet validation set comprising of $312$ scenes. We map the SceneVerse annotations to the 3DSSG vocabulary and construct $1,339$ sub-scenes. The model is trained exclusively on 3DSSG and evaluated on ScanNet without fine-tuning. 
Moreover, since \ours predicts both objects and relations end-to-end, we also evaluate object detection separately on ScanNet~\cite{dai2017scannet}, following the evaluation protocol of SpatialLM~\cite{SpatialLM}. 
Further dataset construction details are provided in Sec.~\ref{supp:implementation_details}.

\PAR{Metrics.}
We evaluate scene graph prediction with standard top-$K$ recall metrics~\cite{lu2016visual}. Recall@$K$ measures whether a ground-truth element is recovered among the top-$K$ ranked predictions. Following prior scene graph protocols~\cite{wald2020learning,wald2022learning,yang2018graph}, we report node recall, predicate recall, and triplet recall. Node recall evaluates object labels, predicate recall evaluates relation labels for matched object pairs, and triplet recall evaluates the full subject--predicate--object prediction. For object detection, we report the IoU$_{0.25}$ F1 metric used by SpatialLM~\cite{SpatialLM}, with more details discussed in Sec.~\ref{supp:implementation_details}.

\PAR{Implementation details.} 
Most prior 3D scene graph methods assume ground-truth object segments or nodes as input during evaluation. In contrast, \ours jointly detects all objects, classifies them, and predicts their relationships during scene graph generation. We therefore establish a one-to-one correspondence between predicted and ground-truth objects via Hungarian matching based on 3D IoU, in order to compute the metrics and evaluate our method. The full details of this matching procedure, as well as the metric computations, are thoroughly discussed in Sec.~\ref{supp:implementation_details}. In addition, all architectural choices, hyperparameters, and training schedules are provided in Sec.~\ref{supp:implementation_details}.

\PAR{Baselines.}
We compare against current state-of-the-art open-vocabulary 3D scene graph methods. ConceptGraphs~\cite{gu2024conceptgraphs} constructs object-level 3D graphs from posed RGB-D sequences by detecting objects in 2D, associating them across views, and using large vision-language models to caption objects and infer inter-object relations. Open3DSG~\cite{koch2024open3dsg} operates on point-cloud object instances, distills 2D vision-language features into a 3D graph neural network, and uses an LLM to generate open-set relationship descriptions. RelationField~\cite{koch2025relationfield} represents a scene as a radiance field and learns scene-specific open-vocabulary object and relationship feature fields by distilling relationship knowledge from multimodal LLMs. ReLaGS~\cite{xie2026relags} operates on 3D Gaussian splats, constructing language-aligned Gaussian object representations and applying GNN-based relational reasoning to predict open-vocabulary relations. 
For each baseline, we follow its suggested evaluation protocol, which for most methods includes providing ground-truth object instances as input, a privilege not available to \ours, which must first detect all objects. To isolate relational reasoning under the same setting, we additionally report \ours$^\dagger$, a variant of our model in which the ground-truth object list is provided in the input prompt and the model predicts only the relationships between the given objects, without retraining.
In addition, since our method relies on an LLM backbone, we also evaluate general-purpose vision-language models without task-specific fine-tuning, namely Qwen3-VL-32B~\cite{bai2025qwen3vltechnicalreport} and GPT-5.4~\cite{singh2025openaigpt5systemcard}. Both are prompted to produce scene graphs in the same JSON format as \ours, using input modality combinations common among recent methods. Further prompting details are provided in Sec.~\ref{supp:implementation_details}.

\begin{table*}[t!]
\centering
\small
\setlength{\tabcolsep}{3.1pt}
\caption{
\textbf{Scene Graph Evaluation on RIO10.}
Recall-based evaluation across object class, predicate, and relationship
triplet prediction. Our \ours achieves the best overall performance.
Notably, RelationField, ReLaGS, and GEODE leverage
\textit{ground-truth object boxes} as input during inference, whereas our method
must jointly detect all objects before predicting their classes and
relationships. \ours$^\dagger$ denotes our variant supplied with ground-truth objects at inference. For a
fair comparison, we report results using point cloud input only during inference, after multi-modal
training. Experiments with different input modality configurations
are provided in~\Cref{tab:multi_modal_rio10}.
}
\vspace{-1mm}

\begin{tabular}{@{}l l c c cc cc cc@{}}
\toprule
\multirow{2}{*}{Model}
& \multirow{2}{*}{\begin{tabular}{c}
Modality\\
inference
\end{tabular}}
& \multirow{2}{*}{\begin{tabular}{c}
GT-free\\
inference
\end{tabular}}
& \multirow{2}{*}{\begin{tabular}{c}
Runtime\\
per scene
\end{tabular}}
& \multicolumn{2}{c}{Object}
& \multicolumn{2}{c}{Predicate}
& \multicolumn{2}{c}{Triplet} \\

\cmidrule(lr){5-6}
\cmidrule(lr){7-8}
\cmidrule(lr){9-10}

& & &
& R@5 & R@10
& R@3 & R@5
& R@50 & R@100 \\

\midrule

Qwen3-VL-32B~\cite{bai2025qwen3vltechnicalreport}
& PCD
& \cmark
& \textcolor{mygreen}{\textasciitilde secs}
& 0.26 & 0.31
& 0.58 & 0.59
& 0.42 & 0.45 \\

GPT-5.4~\cite{achiam2023gpt}
& PCD
& \cmark
& \textcolor{mygreen}{\textasciitilde secs}
& 0.37 & 0.45
& 0.61 & 0.62
& 0.50 & 0.55 \\

GPT-5.4~\cite{achiam2023gpt}
& RGB
& \cmark
& \textcolor{mygreen}{\textasciitilde secs}
& 0.34 & 0.42
& 0.74 & 0.75
& 0.61 & 0.65 \\

GPT-5.4~\cite{achiam2023gpt}
& PCD + RGB
& \cmark
& \textcolor{mygreen}{\textasciitilde secs}
& 0.36 & 0.47
& 0.78 & 0.79
& 0.63 & 0.66 \\

\midrule

ConceptGraphs~\cite{gu2024conceptgraphs}
& RGB-D
& \cmark
& \textcolor{red}{\textasciitilde hours}
& 0.37 & 0.46
& 0.74 & 0.79
& 0.69 & 0.71 \\

Open3DSG~\cite{koch2024open3dsg}
& RGB-D + PCD
& \xmark
& \textcolor{darkorangeyellow}{\textasciitilde mins}
& 0.56 & 0.61
& 0.58 & 0.65
& 0.55 & 0.56 \\

RelationField~\cite{koch2025relationfield}
& NeRF
& \xmark
& \textcolor{red}{\textasciitilde hours}
& \textbf{0.69} & \textbf{0.80}
& 0.76 & 0.82
& 0.73 & 0.74 \\

ReLaGS~\cite{xie2026relags}
& 3DGS
& \xmark
& \textcolor{darkorangeyellow}{\textasciitilde mins}
& 0.68 & 0.79
& 0.79 & \textbf{0.87}
& --- & --- \\

GEODE~\cite{feng2026geode}
& PCD
& \xmark
& \textcolor{darkorangeyellow}{\textasciitilde mins}
& --- & ---
& 0.79 & 0.83
& \textbf{0.76} & 0.80 \\

\midrule

\ours$^\dagger$
& PCD
& \xmark
& \textcolor{mygreen}{\textasciitilde secs}
& --- & ---
& 0.83 & 0.85
& \textbf{0.76} & \textbf{0.92} \\

\ours (ours)
& PCD
& \cmark
& \textcolor{mygreen}{\textasciitilde secs}
& \textbf{0.69} & 0.76
& \textbf{0.85} & \textbf{0.87}
& 0.74 & 0.79 \\

\bottomrule
\end{tabular}

\label{tab:sg_rio10_main}
\end{table*}

\begin{table*}[t]
\centering
\small
\caption{
\textbf{Scene Graph Evaluation on 3DSSG and ScanNet.}
Recall across object class, predicate, and relationship triplet on
the full 3DSSG and ScanNet evaluation sets. On ScanNet, both methods are
evaluated zero-shot after training exclusively on 3DSSG under the same
evaluation protocol. \ours demonstrates stronger zero-shot
predicate generalization. Our method achieves the best overall performance,
confirming the findings from Tab.~\ref{tab:sg_rio10_main}.
}
\vspace{-1mm}
\setlength{\tabcolsep}{4pt}

\begin{tabular*}{\linewidth}{
@{\extracolsep{\fill}} l l @{\hspace{3em}} cc cc cc @{}
}
\toprule
\multirow{2}{*}{Model}
& \multirow{2}{*}{Modality}
& \multicolumn{2}{c}{Object}
& \multicolumn{2}{c}{Predicate}
& \multicolumn{2}{c}{Triplet} \\

\cmidrule(lr){3-4}
\cmidrule(lr){5-6}
\cmidrule(lr){7-8}

& 
& R@5 & R@10
& R@3 & R@5
& R@50 & R@100 \\

\midrule
\multicolumn{2}{c}{\textit{3DSSG (in-domain)}} & & & & & & \\
\cmidrule(lr){1-2}

Qwen3-VL-32B~\cite{bai2025qwen3vltechnicalreport}
& PCD
& 0.12 & 0.14
& 0.45 & 0.46
& 0.35 & 0.36 \\

GPT-5.4
& PCD
& 0.31 & 0.31
& 0.49 & 0.50
& 0.40 & 0.44 \\

\addlinespace[1mm]

Open3DSG~\cite{koch2024open3dsg}
& PCD+RGB-D
& 0.57 & \textbf{0.68}
& 0.63 & 0.70
& 0.64 & 0.66 \\

\ours (ours)
& PCD
& \textbf{0.60} & 0.64
& \textbf{0.82} & \textbf{0.83}
& \textbf{0.68} & \textbf{0.71} \\

\midrule
\multicolumn{2}{c}{\textit{ScanNet (zero-shot)}} & & & & & & \\
\cmidrule(lr){1-2}

Open3DSG~\cite{koch2024open3dsg}
& PCD+RGB-D
& 0.70 & \textbf{0.80}
& 0.51 & 0.53
& 0.71 & \textbf{0.85} \\

\ours (ours)
& PCD
& \textbf{0.74} & 0.77
& \textbf{0.90} & \textbf{0.92}
& \textbf{0.73} & 0.84 \\

\bottomrule
\end{tabular*}
\label{tab:sg_3dssg_full}
\end{table*}

\subsection{Main results}

The primary experimental results for 3D scene graph generation are presented in Tab.~\ref{tab:sg_rio10_main}, evaluated on the RIO10 benchmark using the established recall-based evaluation across object class, predicate, and relationship triplet prediction. \ours achieves the best overall performance, which is particularly notable given the more challenging operating conditions. Unlike the main competing approaches, \ours does not leverage ground-truth objects and their boxes during inference. Therefore, it must jointly detect all objects as graph nodes before predicting their semantic classes and relationships. 
Object classification is therefore the greatest challenge for our method, since competing approaches only classify objects that are already given. Providing ground-truth objects to our model (\ours$^\dagger$) leaves predicate recall roughly unchanged but substantially increases triplet recall, showing that object detection and classification are a major bottleneck for end-to-end triplet prediction. Every predicted relation is anchored to a predicted object, so a missed or misclassified object counts all of its relations as incorrect at the triplet level.
For a fair comparison, we report results using point cloud input only during inference, after multi-modal training. Experiments with different input modality configurations are provided in Sec.~\ref{sec:ablation_analysis}. Beyond performance, our method operates as a fully end-to-end latent method that directly generates the scene graph as a structured JSON, resulting in substantially higher inference speed compared to other multi-stage pipeline alternatives. We also include GPT-5.4 in a frozen end-to-end fashion, as a reference point, evaluated by feeding it several combinations of popular modalities. Although it produces a valid scene graph, its performance falls well short of that of our specialized LLM, underscoring the value of task-specific adaptation.

\begin{figure}[t]
    \centering
    \includegraphics[width=0.97\textwidth]{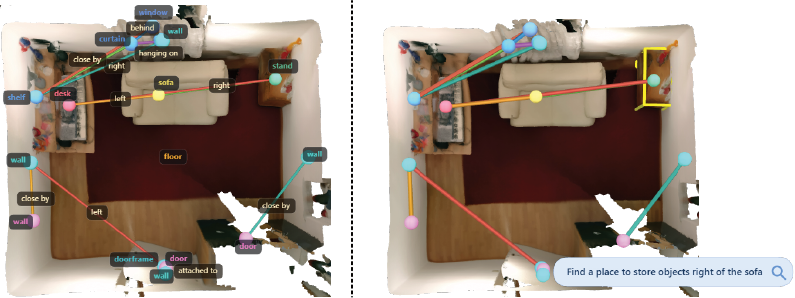}
    \caption{
    \textbf{Qualitative Scene Graph Generation and Open-Vocabulary Querying.} 
    (a) Our \ours generates the 3D scene graph directly from a raw point cloud input, with nodes representing detected objects and edges encoding their predicted semantic relationships. 
    (b) The resulting graph supports open-vocabulary querying, where a natural-language query is matched against the predicted object and relationship labels to retrieve the corresponding elements within the 3D scene.
    }
    \label{fig:demo}
\end{figure}

Experiments on the full evaluation set of $157$ 3DSSG scenes, comprising $548$ subsets, are reported in Tab.~\ref{tab:sg_3dssg_full}. Our method achieves the best overall performance, confirming the findings from Tab.~\ref{tab:sg_rio10_main}. Due to reproducibility constraints and missing open-source implementations of baselines, this comparison is limited to baselines for which results were reported in the original publications or whose code is publicly available. Additionally, since SceneSplat++~\cite{ma2025scenesplatpp} does not provide 3DGS reconstructions for a substantial portion of the 3DSSG dataset, this experiment uses only point cloud inputs during both training and evaluation. 
Moreover, Tab.~\ref{tab:sg_3dssg_full} evaluates zero-shot domain shift, where both methods are trained exclusively on 3DSSG and evaluated on ScanNet. Our method retains strong performance under this domain shift, demonstrating that our framework generalizes well beyond its training dataset.
Furthermore, Fig.~\ref{fig:demo} provides a qualitative illustration of a scene graph generated by \ours on an evaluation scene, together with a demonstration of open-vocabulary querying over the resulting graph. Additional qualitative results with full JSON graph scripts can be found in Fig.~\ref{fig:supp2}.

\subsection{Ablation and Analysis}
\label{sec:ablation_analysis}

Tab.~\ref{tab:multi_modal_rio10} presents an evaluation of all inference-time input configurations of our \ours following multi-modal training on both point clouds and 3DGS. Point-cloud-only inference achieves the strongest performance, which we attribute to the noticeably lower reconstruction quality of the 3DGS counterparts sourced from SceneSplat++~\cite{ma2025scenesplatpp}, introducing noise into the Gaussian Splat encoding branch and reinforcing modality bias~\cite{zheng2025reducing,zheng2025mllmsdeeplyaffectedmodality} toward the more informative point cloud signal. Further discussion and visual illustrations of these quality degradations are provided in Sec.~\ref{supp:addit_exp}. This effect is most pronounced on object detection metrics, where 3DGS-only inference degrades the most significantly, likely due to the difficulty of reliably detecting smaller objects under degraded reconstruction quality. A variant trained and evaluated exclusively on point clouds performs similarly, consistent with the same underlying factors. Nevertheless, we emphasize that multi-modal support is motivated by versatility rather than peak single-modality performance. In practice, different real-world capture pipelines produce different 3D representations, and training with multiple modalities from diverse sources mirrors this setting, enabling a single model with one set of weights to handle whichever representation is available at inference time.

\begin{table}[t]
\centering
\small
\caption{
\textbf{Multi-Modal Analysis on RIO10.} Evaluation of all inference-time input configurations following multi-modal training. Point-cloud-only inference achieves the strongest performance, which we attribute to the noticeably lower reconstruction quality of the 3DGS counterparts (see~\Cref{supp:addit_exp}), reinforcing modality bias toward the more informative point cloud signal. A point-cloud-only trained variant performs similarly for the same reasons.
}
\vspace{1ex}

\begin{tabular}{cc|cc cc cc cc}
\toprule
\multicolumn{2}{c}{Training modality}
& \multicolumn{2}{c}{Inference modality}
& \multicolumn{2}{c}{Object}
& \multicolumn{2}{c}{Predicate}
& \multicolumn{2}{c}{Triplet} \\
\cmidrule(lr){1-2}
\cmidrule(lr){3-4}
\cmidrule(lr){5-6}
\cmidrule(lr){7-8}
\cmidrule(lr){9-10}
PCD & 3DGS
& PCD & 3DGS
& R@5 & R@10
& R@3 & R@5
& R@50 & R@100 \\
\midrule
\checkmark & \checkmark & \checkmark & \checkmark & 0.65 & 0.71 & 0.84 & 0.85 & 0.72 & 0.77 \\
\checkmark & \checkmark & $\times$   & \checkmark & 0.50 & 0.55 & 0.85 & 0.86 & 0.66 & 0.73 \\
\checkmark & \checkmark & \checkmark & $\times$   & 0.69 & 0.76 & 0.85 & 0.87 & 0.74 & 0.79 \\
\midrule
\checkmark & $\times$   & \checkmark & $\times$   & 0.70 & 0.73 & 0.85 & 0.88 & 0.75 & 0.80 \\
\bottomrule
\end{tabular}
\label{tab:multi_modal_rio10}
\end{table}

Tab.~\ref{tab:ablation_rio10} presents an ablation study over the contrastive loss and token fusion, as well as the individual alignment loss components. The results clearly demonstrate that the contrastive loss is critical, as next-token prediction loss alone is insufficient to align the Sonata and Chorus feature spaces for the scene graph generation task. Among the contrastive loss variants, the asymmetric formulation, which pulls Chorus representations toward those of Sonata rather than applying symmetric gradients on both sides, consistently outperforms its symmetric counterpart. This is expected, as Sonata is already well-aligned to the LLM through SpatialLM pretraining, making it the more stable anchor. The asymmetric loss consequently concentrates gradient signal on adapting Chorus to this established shared space. Regarding token fusion, we experimented with several mechanisms for combining spatially co-located features across modalities, finding that simple average pooling yields the best overall performance, suggesting that more complex fusion schemes introduce unnecessary optimization difficulty without a corresponding benefit. 
We further perform leave-one-out ablations of the InfoNCE, cosine similarity, and MSE terms in the alignment objective under the 3DGS-only input setting. Removing any loss term degrades results on nearly every metric, confirming that the three terms provide complementary supervision. InfoNCE has the largest overall impact, with its removal producing the largest drop on metrics, particularly affecting predicate and triplet recall. Cosine similarity is comparable and is slightly more influential for Object R@10, where its removal causes a larger decrease compared with InfoNCE, suggesting that it aids fine-grained object-level alignment. MSE contributes the least overall, leaving predicate recall and Triplet R@100 unchanged when removed, while still improving object-level recall.
Furthermore, visualizations of encoder feature maps before and after contrastive alignment are provided in Fig.~\ref{fig:supp_voxel_features}.

\begin{table*}[t]
\centering
\small
\caption{
\textbf{Ablation Study on RIO10.}
(a) Ablation over contrastive loss formulation and token fusion.
(b) Leave-one-out ablation of the individual alignment loss terms under
the 3DGS-only input setting.
}
\setlength{\tabcolsep}{8pt}
\begin{tabular}{@{}l l@{\hspace{\dimexpr 2\tabcolsep + 7em\relax}} cc cc cc@{}}
\toprule
\multirow{2}{*}{\begin{tabular}{c}
Contrastive \\
loss
\end{tabular}}
& \multirow{2}{*}{Fusion}
& \multicolumn{2}{c}{Object}
& \multicolumn{2}{c}{Predicate}
& \multicolumn{2}{c}{Triplet} \\

\cmidrule(lr){3-4}
\cmidrule(lr){5-6}
\cmidrule(lr){7-8}

&
& R@5 & R@10
& R@3 & R@5
& R@50 & R@100 \\

\midrule
\multicolumn{2}{c}{\textit{(a) Contrastive loss and fusion strategy}} & & & & & & \\
\cmidrule(lr){1-2}
\addlinespace[2pt]

Without
&
Concat.
& 0.60 & 0.64
& 0.80 & 0.81
& 0.65 & 0.74 \\

Without
&
Average
& 0.62 & 0.70
& 0.80 & 0.81
& 0.68 & 0.75 \\

Symmetric
&
Average
& 0.50 & 0.59
& 0.76 & 0.78
& 0.60 & 0.71 \\

Asymmetric
&
Attention
& 0.62 & 0.70
& 0.82 & 0.83
& 0.68 & 0.76 \\

Asymmetric
&
Gated
& 0.53 & 0.59
& 0.79 & 0.81
& 0.62 & 0.72 \\

Asymmetric
&
Average
& 0.65 & 0.71
& 0.84 & 0.85
& 0.72 & 0.77 \\

\midrule
\multicolumn{2}{c}{\textit{(b) Alignment loss design (3DGS-only input)}} & & & & & & \\
\cmidrule(lr){1-2}
\addlinespace[2pt]

\multicolumn{2}{@{}l}{Full-contrastive loss}
& 0.50 & 0.55
& 0.85 & 0.86
& 0.66 & 0.73 \\

\multicolumn{2}{@{}l}{No cosine loss}
& 0.44 & 0.49
& 0.81 & 0.83
& 0.63 & 0.72 \\

\multicolumn{2}{@{}l}{No InfoNCE}
& 0.43 & 0.50
& 0.77 & 0.78
& 0.57 & 0.66 \\

\multicolumn{2}{@{}l}{No MSE}
& 0.45 & 0.54
& 0.85 & 0.86
& 0.62 & 0.73 \\

\bottomrule
\end{tabular}
\label{tab:ablation_rio10}
\end{table*}

We additionally analyze the reliability and scaling of autoregressive JSON generation. Across the complete 3DSSG evaluation set, strict JSON parsing succeeds in $100\%$ of cases whenever the generated output fits within the model's context window. The only observed failure mode is hard output truncation for very large scenes, occurring roughly beyond 40 objects and 1,600 relationships due to the 8k-token context window inherited from the Qwen2.5-0.5B backbone, rather than malformed JSON or syntax collapse. This reveals output length as the primary scaling limitation of the current model. While larger backbones and context windows could improve scalability to denser scenes, we deliberately adopt the compact 0.5B backbone to target resource-constrained embodied robotic platforms. We discuss this trade-off and the resulting scaling limitations in more detail in Sec~\ref{supp:limits}.

\subsection{Synergy Effects for Object Understanding}

Object semantics are often influenced by relational context, so we investigate
whether joint relational supervision in turn benefits object detection and understanding. We
evaluate the point-cloud-only variant of \ours on ScanNet20, with results
reported in Tab.~\ref{tab:obj_det_scannet20}. Joint supervision on object
detection and relational prediction yields measurably stronger detection than
object supervision alone, with a clear margin over SpatialLM. This suggests
that relational context provides an inductive signal for grounding and
localizing individual objects, producing more discriminative and spatially
precise representations than detection supervision alone.

\begin{table}[t]
\centering
\small
\caption{
\textbf{Object Detection Evaluation on ScanNet20.} Our \ours, jointly supervised on both object detection and relationship prediction, notably outperforms SpatialLM, demonstrating that relational supervision provides a useful learning signal for object grounding and localization.
}
\vspace{1ex}
\begin{adjustbox}{max width=\textwidth}
\begin{tabular}{l|c|ccccccccccccccccc}
\toprule
Method 
& \rothead{average}
& \rothead{bathtub} & \rothead{bed} & \rothead{bookshelf} & \rothead{cabinet}
& \rothead{chair} & \rothead{counter} & \rothead{curtain} & \rothead{desk}
& \rothead{door} & \rothead{picture} & \rothead{refrige.}
& \rothead{s.curtain} & \rothead{sink} & \rothead{sofa} & \rothead{table}
& \rothead{toilet} & \rothead{window} \\
\midrule
V-DETR~\cite{shen2024vdetr}
& 65.7 
& 77.4 & 80.1 & 48.8 & 44.4 & 82.5 & \textbf{55.6} & 66.1 
& 68.0 & 62.3 & \textbf{47.4} & 50.8 & 74.0 
& \textbf{79.5} & 72.7 & 55.3 & \textbf{98.0} & 54.9 \\

SceneScript~\cite{avetisyan2024scenescript}
& 49.5 
& 64.5 & 71.1 & 39.7 & 30.3 & 81.1 & 43.4 & 30.9 
& 53.4 & 41.7 & 11.8 & 28.2 & 58.6 
& 48.9 & 68.3 & 55.8 & 77.5 & 36.5 \\

\midrule
SpatialLM~\cite{SpatialLM}
& 66.2 
& 80.6 & 79.9 & 52.0 & 40.0 & \textbf{86.6} & 51.0 & 66.8
& 62.8 & \textbf{67.1} & 29.7 & \textbf{53.6} & \textbf{81.6} 
& 70.7 & 78.9 & 63.9 & 95.4 & 64.3 \\

\ours (ours)
&  \textbf{68.9}
& \textbf{88.5} & \textbf{88.5} & \textbf{56.7} & \textbf{52.4} & 85.7 & 47.6 & \textbf{68.0} 
& \textbf{70.4} & 65.5 & 33.7 & 53.5 & 70.8 
& 73.5 & \textbf{82.5} & \textbf{68.3} & 96.3 & \textbf{69.3} \\
\bottomrule
\end{tabular}
\end{adjustbox}
\label{tab:obj_det_scannet20}
\end{table}

\section{Conclusion}
\label{sec:conclusion}

We presented \ours, an end-to-end method that accepts a 3D point cloud, or Gaussian Splats, or both, and directly outputs a complete scene graph in structured JSON format. Unlike prior work, it requires no explicit intermediate representations, no ground-truth object annotations at inference time, and no proprietary model usage. Motivated purely by versatility, \ours is trained with multi-modal inputs but can be deployed with any single modality at inference, allowing one set of weights to handle diverse inputs. On the 3DSSG benchmark, it achieves state-of-the-art performance on object, predicate, and triplet recall, outperforming methods that assume ground-truth annotations, and our ablations and additional analysis validate each key design choice.
Furthermore, we see broader opportunities beyond pure scene graph generation. Structured relational outputs in text format could ground 3D vision-language models more deeply in physical space, pushing them beyond text-in text-out visual question answering toward richer and more reliable spatially grounded reasoning. Such spatially intelligent systems, in turn, are a critical building block for the next wave of embodied intelligence, enabling autonomous robots to reason about object relations and plan actions, and augmented reality systems to anchor virtual content coherently in the physical world.

\begin{ack}
This research was partially funded by the Ministry of Education and Science of Bulgaria (support for INSAIT, part of the Bulgarian National Roadmap for Research Infrastructure).
Sayan Deb Sarkar is partly supported by the Stanford Doerr School of Sustainability.
\end{ack}

\bibliographystyle{plain}
\bibliography{main}

\clearpage
\appendix

\section{Additional Implementation Details}
\label{supp:implementation_details}

\PAR{End-to-end evaluation protocol.}
Most prior 3D scene graph methods assume ground-truth object segments or ground-truth object nodes as input, so predicted labels are already defined over the correct object set. In contrast, \ours predicts the complete scene graph end-to-end, including both objects and relations. We therefore first establish a one-to-one correspondence between predicted and ground-truth objects using Hungarian matching based on 3D IoU. For rare cases in which a ground-truth object remains unmatched because all remaining predicted objects have zero IoU with it, we use centroid distance as a metric instead of 3D IoU for Hungarian matching. Ground-truth objects that remain unmatched are assigned rank $\infty$ and therefore cannot contribute to recall at any finite $K$. Since \ours decodes free-form text labels rather than closed-set class logits, we compute ranks by semantic comparison against the evaluation vocabulary, following~\cite{koch2024open3dsg}. For node recall, the predicted object label is compared to all object classes in the 3DSSG vocabulary using Jina~\cite{gunther2023jina} embeddings. For predicate recall, predicted relation labels between matched object pairs are ranked using SBERT embeddings~\cite{reimers2019sentencebert}. Triplet recall is stricter: a ground-truth triplet is recovered only if both endpoint objects are matched and the subject, predicate, and object labels are semantically recovered within the top-$K$ ranking.

\begin{figure}[h]
\centering
\begin{minipage}{0.82\linewidth}
\includegraphics[width=\linewidth]{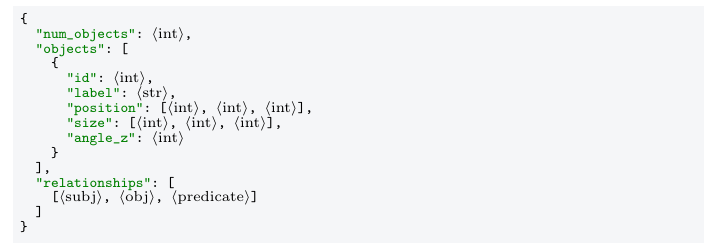}
\end{minipage}
\vspace{-2mm}
\caption{\textbf{Scene Graph JSON Template.} \ours decodes object attributes and pairwise relations as a structured JSON scene graph.}
\label{fig:json_schema}
\vspace{-3mm}
\end{figure}

\begin{figure}[t]
    \centering
    \includegraphics[width=\textwidth]{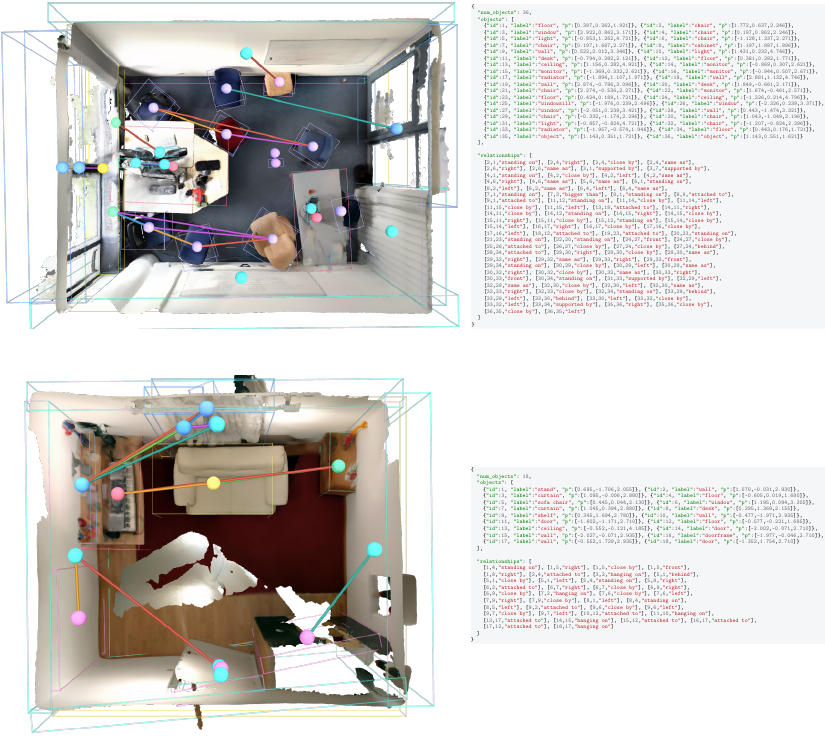}
    \vspace{-2mm}
    \caption{\textbf{\ours Scene Graph Prediction.}
We show results of \ours scene graph prediction on two scenes, given the input 3D scene and the text prompt. Each row shows an input 3D scene with predicted object boxes and relations on the left, and the corresponding generated JSON scene graph on the right.}
    \label{fig:supp2}
    \vspace{-3mm}
\end{figure}

\PAR{Scene Graph Prompting Format.}
The decoder is prompted to generate the full scene graph directly as JSON. The requested response follows the template shown in Figure~\ref{fig:json_schema}. Following SpatialLM~\cite{SpatialLM}, we represent spatial quantities as integer tokens in the language sequence. Coordinates are quantized into $1{,}280$ bins with $2.5$cm resolution, and the LLM predicts these discrete values as part of the JSON output. During post-processing, the predicted integers are mapped back to continuous metric values using the inverse quantization and normalization transform, restoring positions and box sizes to the original scene coordinate system. This formulation lets the model express objects, geometry, and relations within a single language-decoding interface. Figure~\ref{fig:supp2} qualitatively shows the output from \ours given the prompt and the input 3D scene.

\PAR{VLM Baseline Implementation Details.}
We implement the zero-shot VLM baselines using a common prompting and parsing pipeline. For Qwen, we run Qwen3-VL-32B-Instruct locally on a single NVIDIA H200 GPU and evaluate it with the colored PLY point-cloud input. For GPT-5.4, we obtain predictions through API calls, which allows us to evaluate longer multi-modal prompts containing both the serialized point cloud and selected RGB views. Each validation scene is converted into a prompt asking the model to produce a single JSON scene graph with the same template as \ours. We evaluate several input variants: a compact text serialization of the colored PLY point cloud, RGB views and combined PLY+RGB inputs. For the RGB variants, we select up to $12$ views per scene from the parent 3RScan sequence. To choose informative views, we project the split point-cloud centroid and bounding-box corners into the available camera frames, score candidate views by how well they cover the target scene region, and greedily select a diverse set of high-coverage views. The PLY input provides a compact geometric representation of the scene, serialized as metric 3D coordinates with RGB values.

\PAR{ScanNet Object Detection Comparison.}
To make the object-detection comparison with SpatialLM controlled at the evaluation level, we follow the SpatialLM ScanNet protocol and evaluate only the decoded object nodes, i.e., class labels and 3D boxes, while ignoring predicted relationships at test time. Starting from the same SpatialLM checkpoint, we fine-tune \ours on SceneVerse using their scene-graph annotations and the same JSON decoding format as in the main scene-graph task. This fine-tuning exposes the model to complete scene-graph supervision, including both object and relationship tokens, but the ScanNet metric is computed solely from the object portion of the output. Therefore, Tab.~5 should be interpreted as measuring whether our end-to-end scene-graph fine-tuning preserves or improves object detection under the SpatialLM evaluation protocol, rather than as a causal ablation isolating relational supervision from all other factors such as modality mixture, Chorus parameters, or the auxiliary alignment loss. During the evaluation of our model in this setup, we follow the evaluation approach from~\cite{SpatialLM}.

\paragraph{ScanNet Dataset Construction.}
We map the SceneVerse~\cite{jia2024sceneverse} object and relationship annotations to the 3DSSG vocabulary using normalized label matching and manually defined aliases, following the approach from~\cite{wu2021scenegraphfusion}. Because complete ScanNet scenes generally contain more objects than the 3DSSG training examples, we divide them into local sub-scenes containing five to nine objects. The largest floor instance is included in every sub-scene as a common spatial reference, while each remaining object is assigned to exactly one sub-scene. Objects are grouped according to spatial proximity and neighborhood connectivity, where two objects are considered neighbors if their axis-aligned bounding boxes overlap after expansion by $0.5\,\mathrm{m}$. Using a fixed random seed for reproducibility, we grow each group by adding the neighboring object closest to its centroid in the horizontal plane. When no unassigned neighbor remains, we instead add the closest unassigned object. Each resulting point cloud contains only the selected object instances, and its graph retains relationships whose endpoints are both present. This procedure produces $1,339$ sub-scenes from the $312$ ScanNet validation scenes.

\PAR{Architecture Details.}
We use Qwen2.5-0.5B~\cite{qwen2025qwen25technicalreport} as the language decoder, Sonata~\cite{wu2025sonata} for point-cloud encoding, and Chorus~\cite{li2025chorus} for Gaussian-splat encoding. Per-voxel 3D features are projected into the language-model embedding space using a lightweight MLP, following SpatialLM~\cite{SpatialLM}. The Sonata encoder, projector, and language decoder are initialized from the released SpatialLM checkpoint. We fully fine-tune the 3D encoders, projector, and decoder, and use modality dropout during training so that a single model supports point-cloud-only, Gaussian-only, and fused inference. %
The Qwen2.5-0.5B language decoder has hidden size $h=896$. The Sonata point-cloud encoder uses a voxel size of $2.5$cm and produces per-voxel features of dimension $d=512$. The Chorus Gaussian-splat encoder uses the same voxel size and output feature dimension as Sonata. Per-voxel features from Sonata and Chorus are projected into the Qwen2.5-0.5B embedding space using a per-token MLP with dimensions $d\rightarrow d \rightarrow h$, following SpatialLM~\cite{SpatialLM}.

\PAR{Optimization Details.}
We fully fine-tune the 3D encoders, projector, and LLM decoder for $5$ epochs. We use AdamW with learning rate $2\times10^{-5}$, an effective batch size of $\mathcal{B}=8$, and a cosine learning-rate schedule with a $3\%$ warmup ratio, following SpatialLM~\cite{SpatialLM}. For multimodal robustness, we apply modality dropout with dropout probabilities of $35\%$ for the point-cloud modality and $15\%$ for the 3DGS modality. We use a two-stage loss schedule for cross-modal alignment. We employ a two-stage training schedule in which the auxiliary cross-modal alignment loss is applied with weight $\alpha=0.25$ during the first half of the training and disabled afterwards. For the InfoNCE loss, we set the temperature to $\tau=0.07$. Training is conducted on one NVIDIA H200 GPU for approximately $4$ hours.

\PAR{Compute Resources.}
All experiments were run on an on-premises institutional GPU cluster. Each training run used a single bare-metal node with one NVIDIA H200 GPU with $143$GB of GPU memory, Intel Xeon Gold 6548Y+ CPUs, and NVMe SSD storage. Each run requested $8$ CPU cores, $80$GB of host memory, and one H200 GPU, with a maximum wall-clock time of $6$ hours. A full fine-tuning run for \ours takes approximately $4$ hours. All reported training configurations use the same compute allocation. Across the reported training runs and ablations, the total compute is approximately $1000$ H200 GPU-hours. Inference and evaluation are lightweight relative to training and are run on the same cluster allocation. The full research project required additional exploratory compute for debugging, preliminary experiments, and ablations beyond the final reported runs; these exploratory runs were not required to reproduce the final experimental results.

\section{Additional Experimental Analysis}
\label{supp:addit_exp}

\begin{figure}[t]
    \centering
    \includegraphics[width=\textwidth]{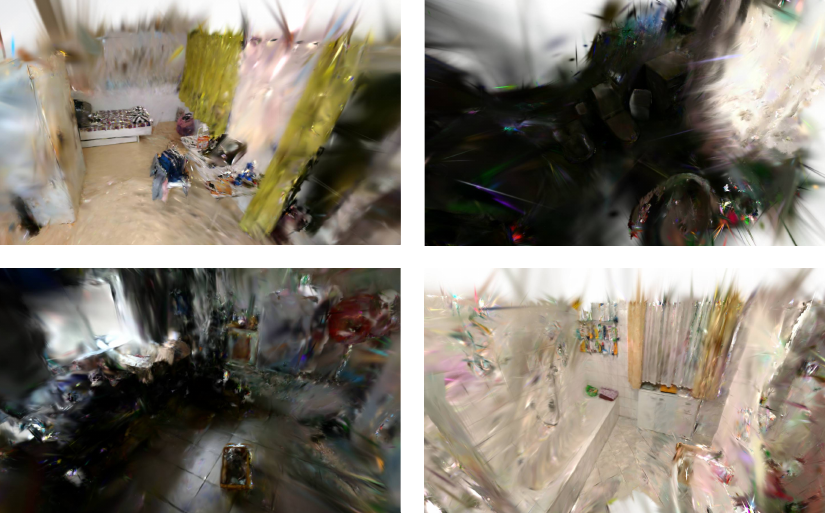}
    \vspace{-2mm}
    \caption{\textbf{Examples of degraded 3DGS reconstructions on RIO10.} Most 3D Gaussian splat reconstructions in RIO10 exhibit substantial artifacts, including blurred geometry, floaters, missing structure, and distortions.}
    \label{fig:supp_bad_3dgs}
    \vspace{-3mm}
\end{figure}

\PAR{3DGS Reconstruction Quality} Table~\ref{tab:multi_modal_rio10} shows that point-cloud-only inference achieves the strongest performance after multi-modal training. We attribute this behavior primarily to the reconstruction quality of the available Gaussian splats. 3D Gaussian Splatting reconstructions for 3DSSG and RIO10  were taken from \cite{ma2025scenesplatpp} and contain substantial artifacts, including blurred geometry, floaters, missing structure, and strong view-dependent distortions. This is depicted in Figure~\ref{fig:supp_bad_3dgs}. As a result, the Gaussian-splat encoder often receives a noisier and less spatially reliable signal than the point-cloud encoder, biasing the model toward the point-cloud modality. This also explains why a point-cloud-only trained variant performs similarly, since in this setting, the additional Gaussian-splat input often does not provide sufficiently reliable complementary information.

\begin{figure}[t]
\centering
\begin{minipage}{0.98\linewidth}
\includegraphics[width=\linewidth]{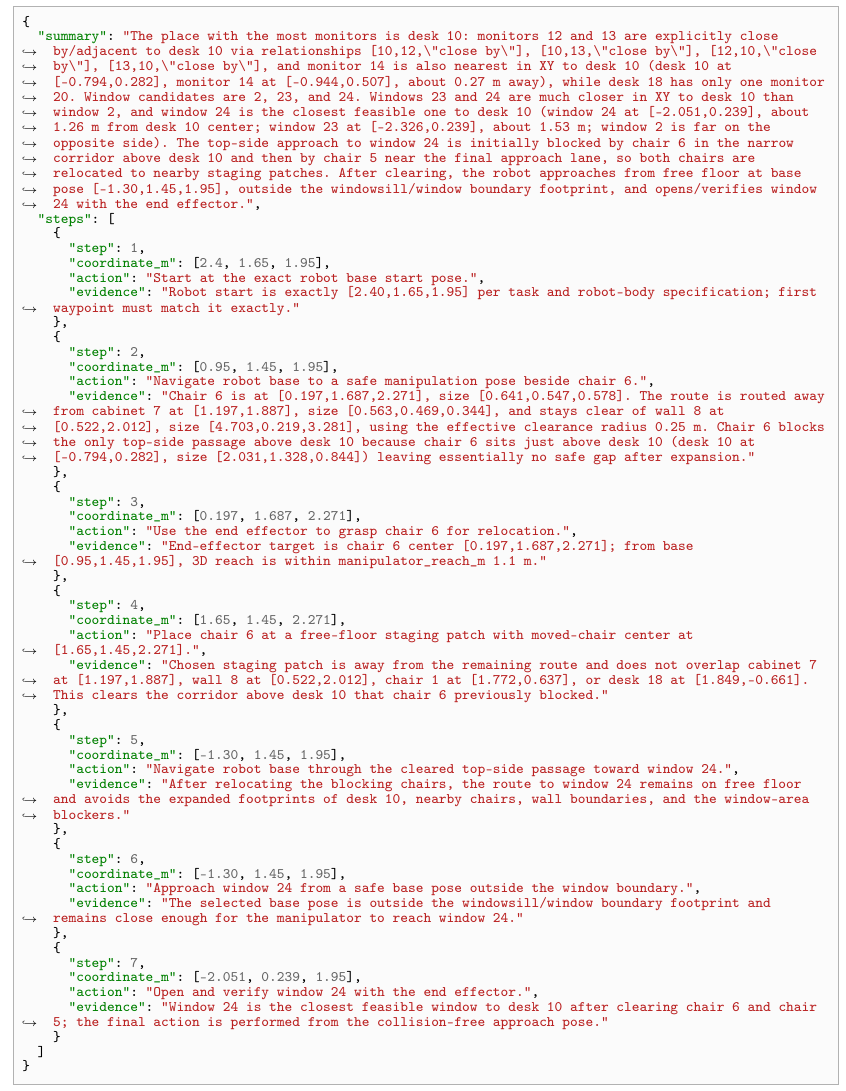}
\end{minipage}
\caption{\textbf{Embodied Planning Using Predicted Scene Graphs as Grounding.}
The JSON trace records the selected target desk, selected window, blocking objects, and the resulting navigation/manipulation sequence.}
\label{fig:vqa_reasoning_json}
\end{figure}

\begin{figure}[t]
    \centering
    \includegraphics[width=\textwidth]{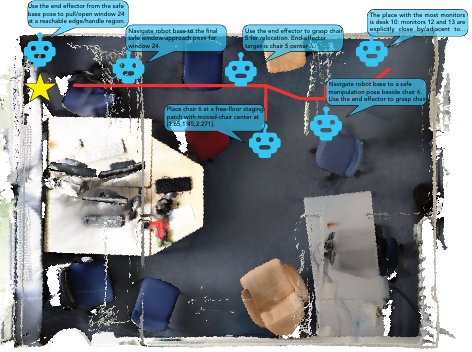}
    \caption{\textbf{Qualitative Embodied Planning Using 3D Scene Graph Reasoning.} Given a natural-language instruction, the predicted scene graph is used to ground the LLM reasoning and identify the desk with the most nearby monitors, select the closest feasible window, and derive a grounded right-to-left navigation plan. The robot first clears two blocking chairs along the upper passage, then approaches the selected window from a safe base pose and performs the final opening/verification action. This example illustrates how explicit object nodes, relations, and 3D geometry support compositional reasoning over the scene.}
    \label{fig:vqa_demo}
\end{figure}

\PAR{Embodied Planning Using 3D Scene Graphs}
We qualitatively demonstrate how the generated scene graph can serve as an explicit grounding representation for embodied planning in 3D scenes. Given a natural-language instruction and the predicted graph, we query GPT-5.4 to identify relevant objects, reason over their semantic relationships, and ground the resulting plan back into metric 3D space. Figure~\ref{fig:vqa_demo} shows an example in which a robot is instructed to navigate from the right side of the room and open the window closest to the desk with the most monitors. The planning model first identifies the monitor-rich workspace by aggregating relations between desks and nearby monitors, then selects the nearest feasible window among the predicted window nodes. It then uses the predicted object geometry to avoid large obstacles and identifies two movable chairs that obstruct the passage. This provides a preliminary downstream probe of the embodied-intelligence use cases discussed in the main paper: structured relational outputs can support spatial reasoning and action sequencing. However, this example is intended only as a qualitative demonstration of how predicted scene graphs may support embodied planning, not as a full solution to robotic navigation and manipulation. A more detailed excerpt from the GPT-5.4 planning response is shown in Figure~\ref{fig:vqa_reasoning_json}.
This example illustrates how the explicit scene-graph representation enables reasoning beyond direct object detection. The model output is not only a list of objects, but a structured representation that supports compositional spatial queries such as counting monitors near desks, comparing candidate windows by distance, and producing a grounded sequence of navigation and manipulation steps. We emphasize that this is a qualitative downstream demonstration rather than a separately trained robotics system. The trajectory and chair-relocation actions are derived from the predicted graph, object locations, and specified robot constraints.

\begin{figure}[h]
    \centering
    \includegraphics[width=0.85\textwidth]{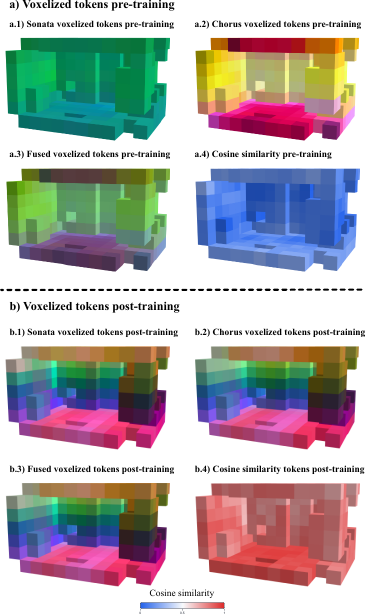}
    \caption{\textbf{Effect of Auxiliary Loss on Per-Voxel Feature Alignment} We visualize Sonata, Chorus, fused per-voxel features, and the corresponding cosine-similarity map between per-voxel features. Prior to training (a), the point-cloud and Gaussian-splat encoders produce inconsistent voxel-level feature patterns. After training, the modality-specific representations become aligned, which is shown by the substantial increase in cosine-similarity across corresponding voxels.}
    \label{fig:supp_voxel_features}
\end{figure}

\PAR{Effect of Auxiliary Loss on Per-Voxel Feature Alignment}
Figure~\ref{fig:supp_voxel_features} illustrates the effect of the auxiliary cross-modal alignment loss on voxel-level representations. We visualize features at the final encoded voxel resolution, where Sonata, Chorus, and fused features are associated with corresponding voxel locations. For feature visualization, we compute a PCA basis from the pooled Sonata, Chorus, and fused voxel features from the same scene. We then map the first three projected coordinates to RGB using normalization across all three feature maps. We additionally report per-voxel cosine similarity between aligned feature vectors, with red denoting higher similarity. Before training, the Sonata and Chorus features are weakly aligned. After training, the modality-specific features become more consistent, suggesting that the auxiliary loss encourages co-located point-cloud and Gaussian splat tokens to align in the shared voxel space.

\section{Comparison to Relevant Methods.}
\label{supp:runtime}

\begin{figure}[t]
    \centering
    \includegraphics[width=\textwidth]{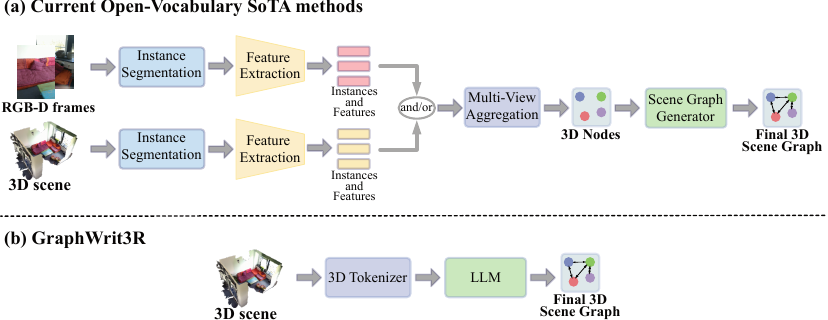}
    \caption{\textbf{Comparison of Open-Vocabulary 3D Scene Graph Generation Pipelines.}
    (a) Recent state-of-the-art approaches typically rely on multi-stage pipelines: 3D scenes and/or RGB-D frames are first processed by instance segmentation and feature extraction modules, followed by feature aggregation and scene graph generation.
    (b) In contrast, \ours directly tokenizes the input 3D scene and uses an LLM to generate the final 3D scene graph in a single forward pass, avoiding explicit intermediate processing stages.}
    \label{fig:supp_highlevel}
\end{figure}

In Table~\ref{tab:param_compare}, we provide an approximate comparison of model size, trainable parameters, and per-scene runtime across open-vocabulary 3D scene graph methods. Since existing methods differ substantially in their use of frozen backbones, external APIs, per-scene optimization, and multi-stage preprocessing, the numbers are intended as order-of-magnitude estimates rather than strict benchmarks. We further provide a high-level comparison between prior open-vocabulary scene graph methods, as shown in Figure~\ref{fig:supp_highlevel}.

\PAR{Parameter Count.}
We define total number of parameteres as the inference-time model footprint required by the method, including frozen foundation models that must be loaded or queried. Trainable parameters denote the number of parameters in the reported training setup. For API-based methods, we report zero trainable parameters, since no method-specific parameters are optimized locally, while the total parameter count reflects the approximate scale of the proprietary model family used during inference.

For \ours, the parameter count includes the Qwen2.5-0.5B decoder, the Sonata point-cloud encoder, the Chorus Gaussian-splat encoder, the SpatialLM projection modules, and the additional fusion parameters. This gives a total of approximately $736.6$M parameters, all of which are trainable in our setup.

\PAR{Runtime.}
Runtime is reported at the granularity of seconds, minutes, or hours per scene. Methods based on per-scene optimization, repeated VLM/LLM querying, or dense 2D--3D association typically incur minute to hour-scale inference costs. In contrast, \ours performs end-to-end scene graph prediction, yielding second-scale inference in our implementation.

\PAR{Input Assumptions.}
We separately indicate whether a method requires ground-truth object nodes or object proposals as input during evaluation. This distinction is important because many scene graph methods evaluate relation prediction conditioned on known objects, whereas \ours predicts both objects and relations without ground-truth nodes at inference time. We also mark methods that rely on proprietary models, since such dependencies affect reproducibility, deployment cost, and the feasibility of fully local evaluation.

\section{Limitations}
\label{supp:limits}
\ours has several limitations. First, our experiments focus on indoor reconstructed scenes, primarily 3DSSG for scene graph prediction and ScanNet for object detection, so robustness to outdoor, dynamic, and building-scale scenes remains untested. Second, the multi-modal setting is limited by the quality and availability of Gaussian-splat reconstruction. In our experiments, point-cloud-only inference performs best, while 3DGS inputs are affected by noisy or incomplete reconstructions. Third, our scene-graph experiments inherit a broader limitation of the 3DSSG benchmark protocol. Existing methods are commonly trained and evaluated on small annotated subgraphs, often containing only around nine object nodes, rather than on complete scene-level graphs. This makes the benchmark useful for controlled comparison, but it also means that reported results may underestimate the difficulty of full-scene graph generation, long-range relational reasoning, and scaling to cluttered environments with many objects. This is also relevant to our autoregressive JSON formulation. In full-scene stress tests, strict JSON parsing succeeds in $100\%$ of cases whenever the output fits within the model's 8k-token context window. The only observed failure mode is hard truncation for very large scenes, roughly beyond $40$ objects and $1,600$ relationships. Thus, scaling is currently limited primarily by output length rather than JSON syntax reliability. Larger backbones or context windows could alleviate this bottleneck, at the cost of increased computational requirements. Finally, because \ours predicts objects and relations end-to-end, evaluation requires Hungarian object matching and semantic text-embedding-based ranking of free-form labels. The resulting scores may therefore depend on these protocol choices.

\begin{table*}[t]
\caption{\textbf{Computational Comparison of Open-Vocabulary 3D Scene Graph Methods.}
We report approximate inference-time model footprint, trainable parameters, per-scene runtime, input assumptions, and reliance on proprietary models. Among the compared methods, \ours is the only fully local approach that simultaneously avoids ground-truth object nodes as inputs at evaluation time and operates at second-scale runtime.}
\centering
\footnotesize
\setlength{\tabcolsep}{4pt}
\begin{tabular}{@{}llcccc@{}}
\toprule
Method
& Total Params
& Trainable Params
& Runtime / Scene
& \begin{tabular}{c}
No GT nodes\\
as inputs
\end{tabular}
& \begin{tabular}{c}
Uses proprietary\\
models 
\end{tabular} \\

\midrule

\ours (ours)
& $\approx 736.6$M
& $\approx 736.6$M
& $\sim$seconds
& \checkmark
& $\times$\\

RelationField~\cite{koch2025relationfield}
& $\approx 200$M
& $\approx 200$M
& $\sim$hours
& $\times$
& $\times$ \\

Open3DSG~\cite{koch2024open3dsg}
& $\sim 8$B
& $\approx 111$M
& $\sim$minutes
& $\times$
& $\times$ \\

ReLaGS~\cite{xie2026relags}
& $\sim 1.8$B
& $\approx 8.2$M
& $\sim$minutes
& $\times$
& $\times$ \\

Beyond Bare Queries~\cite{linok2025beyond}
& $\sim 1$T
& $0$
& $\sim$hours
& $\times$
& \checkmark\\

ConceptGraphs~\cite{gu2024conceptgraphs}
& $\sim 1$T
& $0$
& $\sim$hours
& \checkmark
& \checkmark\\

OpenFunGraph~\cite{zhang2025open}
& $\sim 1$T
& $0$
& $\sim$hours
& \checkmark
& \checkmark \\

FunFact\cite{Fu_2026_funfact}
& $\sim 1$T
& $0$
& $\sim$hours
& \checkmark
& \checkmark\\

FunGraph~\cite{rotondi2025fungraph}
& $\sim 1$T
& $0$
& $\sim$hours
& \checkmark
& \checkmark \\

\bottomrule
\end{tabular}
\label{tab:param_compare}
\end{table*}

\section{Broader Societal Impact}
\label{supp:broader_impact}

\ours aims to improve structured 3D scene understanding by producing explicit object-and-relation scene graphs from point clouds or Gaussian splats. This can benefit applications that require spatially grounded reasoning, such as robotics, augmented reality, assistive navigation, embodied question answering, and scene editing, while reducing reliance on multi-stage pipelines and proprietary APIs. However, stronger 3D scene understanding also introduces risks if deployed without safeguards. Scene graphs inferred from indoor scans may reveal private information about room layouts, object ownership, or activity patterns, and the same capabilities that support robot planning could be misused for surveillance, inventorying private spaces, or unauthorized physical interaction with objects. In embodied settings, incorrect detections or relations may also lead to unsafe plans, such as navigating through obstacles or manipulating the wrong object. Our work is evaluated as a benchmark perception method rather than an autonomous deployment system, but practical use should require consent for 3D data capture, privacy-preserving data handling, domain-specific safety checks, uncertainty-aware planning, and human oversight before physical actions are executed. The model should not be treated as a complete safety-critical robotics system without further validation in the target environment.

\section{Existing Asset Licenses}
\label{supp:existing_asset_licenses}

We use existing assets only for non-commercial academic research and evaluation, and do not redistribute restricted raw datasets, benchmark assets, simulator data, or pretrained checkpoints. All datasets, codebases, pretrained models, and baselines are cited in the main paper and used under their stated licenses or access terms.

We use 3DSSG annotations together with the underlying 3RScan/RIO10 assets, following the 3RScan Terms of Use required by the official 3DSSG release~\cite{wald2020learning, Wald2019RIO}. ScanNet assets are used under the ScanNet Terms of Use, which restrict use to non-commercial research and educational purposes~\cite{dai2017scannet}. SceneVerse code is released under the MIT License, while its data are derived from multiple source datasets and therefore inherit the terms of the corresponding underlying datasets~\cite{jia2024sceneverse}. SceneSplat++/SceneSplat-49K is released through a gated dataset card listing CC-BY-SA-4.0, with additional non-commercial research and educational-use conditions. For re-packaged components, we follow both the derived release terms and the original dataset licenses~\cite{ma2025scenesplatpp,li2025SceneSplat7k}. Sonata code is Apache License 2.0 and Sonata weights are CC-BY-NC 4.0~\cite{wu2025sonata}. SpatialLM1.1-Qwen-0.5B uses CC-BY-NC-4.0 model weights and Apache-2.0 Pointcept-based code~\cite{SpatialLM,wu2024ptv3}. Qwen2.5-0.5B-Instruct is released under Apache License 2.0~\cite{qwen2025qwen25technicalreport}. Chorus~\cite{li2025chorus} code and model weights are released under the CC-BY-SA 4.0 license. For repository-based assets without stable release tags, the exact commit hashes used in our experiments are documented in the released configuration files.

\end{document}